\pdfoutput=1

\documentclass[11pt]{article}

\usepackage[final]{acl}

\usepackage{times}
\usepackage{latexsym}
\usepackage[T1]{fontenc}

\usepackage[utf8]{inputenc}

\usepackage{microtype}

\usepackage{inconsolata}

\usepackage{graphicx}

\usepackage{booktabs}  
\usepackage{graphicx}  
\usepackage{xcolor}    
\usepackage{colortbl}
\usepackage{tabularx}
\usepackage{multirow}
\usepackage{multicol}
\usepackage{amsmath}
\usepackage{amssymb}
\usepackage{mathtools}
\usepackage{amsthm}
\usepackage{subcaption}

\title{Rethinking Diverse Human Preference Learning \\through Principal Component Analysis}

\author{Feng Luo \\
  Rice University \\
  \texttt{fl38@rice.edu} \\\And
  Rui Yang \\
  Affiliation / Address line 1 \\
  \texttt{email@domain} \\
  }

\author{
 \textbf{Feng Luo\textsuperscript{1}\thanks{Equal contribution.}},
 \textbf{Rui Yang\textsuperscript{2}\footnotemark[1]},
 \textbf{Hao Sun\textsuperscript{3}},
 \textbf{Chunyuan Deng\textsuperscript{1}},
 \textbf{Jiarui Yao\textsuperscript{2}},
 \textbf{Jingyan Shen\textsuperscript{4}},
\\
 \textbf{Huan Zhang\textsuperscript{2}},
 \textbf{Hanjie Chen\textsuperscript{1}}
\\
 \textsuperscript{1}Rice University,
 \textsuperscript{2}University of Illinois at Urbana-Champaign,
 \\
 \textsuperscript{3}University of Cambridge,
 \textsuperscript{4} Columbia University
\\
 \small{
   \href{mailto:fl38@rice.edu}{fl38@rice.edu}, \href{mailto:ry21@illinois.edu}{ry21@illinois.edu}, \href{mailto:hanjie@rice.edu}{hanjie@rice.edu}
 }
}

\begin{document}
\maketitle
\begin{abstract}
Understanding human preferences is crucial for improving foundation models and building personalized AI systems. However, preferences are inherently diverse and complex, making it difficult for traditional reward models to capture their full range. While fine-grained preference data can help, collecting it is expensive and hard to scale. In this paper, we introduce Decomposed Reward Models (DRMs), a novel approach that extracts diverse human preferences from binary comparisons without requiring fine-grained annotations. Our key insight is to represent human preferences as vectors and analyze them using Principal Component Analysis (PCA). By constructing a dataset of embedding differences between preferred and rejected responses, DRMs identify orthogonal basis vectors that capture distinct aspects of preference. These decomposed rewards can be flexibly combined to align with different user needs, offering an interpretable and scalable alternative to traditional reward models. We demonstrate that DRMs effectively extract meaningful preference dimensions (e.g., helpfulness, safety, humor) and adapt to new users without additional training. Our results highlight DRMs as a powerful framework for personalized and interpretable LLM alignment. Our code is available at \url{https://github.com/amandaluof/DRMs}
\end{abstract}


\begin{figure*}[t!]
\vspace{-0.6cm}
    \centering
    \includegraphics[width=1.0\textwidth]{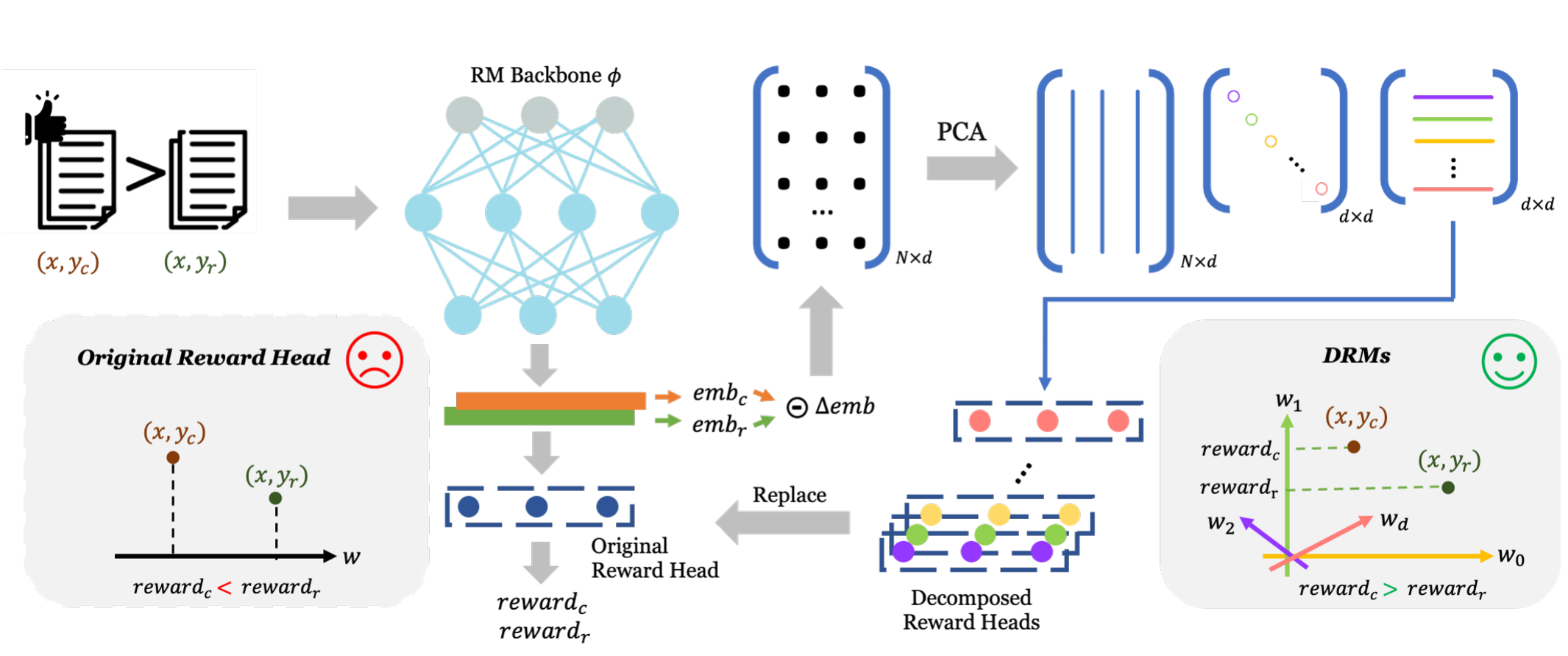}
    \caption{Illustration of the decomposition pipeline in DRMs. In the original single-dimensional head, a prompt–response pair can be predicted incorrectly. In contrast, DRMs capture preferences along multiple dimensions, aligning more effectively with the complex and multifaceted nature of human preferences.}
    \label{fig: Framework.}
\vspace{-0.2cm}
\end{figure*}

\section{Introduction}
Reinforcement Learning from Human Feedback (RLHF) \cite{stiennon2020learning,ouyang2022training,bai2022training,team2023gemini,grattafiori2024llama3herdmodels,liu2024deepseek} has proven to be a powerful approach for fine-tuning large language models (LLMs) to serve as better general assistants \cite{team2023gemini,achiam2023gpt}, AI agents \cite{nakano2021webgpt, zhao2024expel,yang2025embodiedbench}, and clinical decision support systems \cite{tian2023chimed,wang2024assessing,chen2025benchmarking}. Typically, LLMs are optimized using a scalar reward model trained on human preference data, acting as a proxy for overall user satisfaction. However, this approach has two key limitations: (1) it often reflects the preferences of the majority, potentially marginalizing underrepresented groups \cite{chakraborty2024maxmin,chidambaram2024direct} and failing to capture the full diversity of human preferences, and (2) it struggles to represent the complex, multifaceted, and sometimes conflicting nature of human preferences with a single scalar reward \cite{jang2023personalized,rame2024rewarded,yang2024rewards,zhou2024beyond}.

As demand grows for more personalized LLMs \cite{zhang2024personalization}, researchers have explored ways to capture fine-grained, multidimensional human preferences. Some studies have introduced datasets that evaluate multiple aspects such as relevance, correctness, completeness, helpfulness, and harmlessness \cite{wu2023fine,wang2023helpsteer,cui2023ultrafeedback,pitis2024improving}. Building on these datasets, others have proposed multi-objective optimization methods \cite{qiu2024traversing} to accommodate diverse user needs \cite{yang2024rewards,yang2024metaaligner,wang2024conditional}. However, these approaches face significant challenges: collecting fine-grained human annotations is expensive, and using GPT-generated labels can introduce biases. Considering that large-scale binary preference datasets—where users simply compare two responses—are easier to collect and more widely available, we ask the question: 
\emph{\textbf{Can we infer multidimensional human preferences directly from large-scale binary comparisons?}}

To address this question, we propose Decomposed Reward Models (DRMs), a framework that extracts fine-grained human preferences using binary comparison data. Unlike traditional probabilistic models such as Bradley-Terry (BT) \cite{bradley1952rank} or pairwise preference models \cite{jiang2023llm}, our approach represents human preferences as $d$-dimensional vectors, corresponding to the learned weights of the final linear layer in a reward model. We show that this vector-based representation establishes a connection between Principal Component Analysis (PCA) and preference learning. Our method constructs a dataset of embeddings for all prompt-response pairs and applies PCA to the differences between chosen and rejected response embeddings. This process identifies orthogonal basis vectors, each capturing a distinct human preference direction. These basis vectors can be combined with their corresponding feature extractor to construct reward models. The framework is shown in Figure \ref{fig: Framework.}.
DRMs are particularly effective for personalized preference learning. Given a small adaptation dataset from a new user, we compute coefficients for the basis vectors and form a linear combination that best aligns with the user’s preferences. This approach allows for flexible, scalable, and interpretable preference modeling without requiring additional training.

In our experiments, we first show that the learned DRM basis vectors capture diverse human preference attributes. For instance, one vector may strongly correlate with safety, while another may reflect humor. Next, we evaluate DRM’s effectiveness in adapting to user preferences at test time. Our results show that DRMs outperform both the single-head reward model and previous test-time alignment methods based on reward ensembles. These findings emphasize DRM’s advantages in modeling and adapting to diverse human preferences. In addition, we also provide an explainability study to better understand human preferences.

In summary, our work makes three key contributions. (1) We propose a vector-based representation of human preferences, providing a more structured and interpretable approach to preference learning. (2) We establish a connection between preference learning and PCA, enabling the extraction of diverse and meaningful reward components without additional training. (3) We empirically validate DRMs, demonstrating their effectiveness in capturing diverse human preferences and adapting to new users at test time. These findings highlight DRMs as a scalable and flexible alternative to traditional reward models, with promising applications in interpretable AI and LLM personalization.

\section{Preliminary}
The Bradley-Terry (BT) model \cite{bradley1952rank} is a probabilistic framework commonly used in preference learning. Given a prompt \( x \) and two responses, \( y_1 \) and \( y_2 \), the BT model defines the probability of \( y_1 \) being preferred over \( y_2 \) as:  
$$ 
P(y_1 \succ y_2 | x) = \frac{\exp(r(x,y_1))}{\exp(r(x,y_1)) + \exp(r(x,y_2))}  
$$
where \( r(x,y) \) represents the reward model, which assigns a score to a given prompt-response pair.  
Given a dataset of comparisons \(\mathcal{D} = \{(x_i, y_i^c, y_i^r)\}_{i=1}^{N}\), where \( y_i^c \) is the preferred response and \( y_i^r \) is the rejected one, the reward model is trained by maximizing the likelihood of human preferences. This results in the following objective:  
\begin{equation}  
\begin{aligned} \label{eq:bt_loss}
    \max_{\theta} \mathbb{E}_{(x_i, y_i^c, y_i^r) \sim \mathcal{D}} \left[\log \sigma \left( r_\theta(x_i, y_i^c) - r_\theta(x_i, y_i^r) \right) \right]  
\end{aligned}
\end{equation}  
where \( r_\theta(x, y) \) is the reward score parameterized by \( \theta \), and \( \sigma(\cdot) \) denotes the sigmoid function.  
By minimizing this loss, the reward model learns to assign higher scores to human-preferred responses, making it a useful proxy for human preferences.


\section{Methodology}\label{sec:method}

The standard approach to preference learning relies on a scalar-valued reward model, which may not adequately capture the full complexity of human preferences. To address this limitation, we first introduce a vector representation of human preference and establish its connection to PCA. Building on this, we propose a PCA-based method that decomposes human preference into multiple basis vectors, allowing any preference to be represented as a linear combination of these vectors. This approach enables a novel paradigm for diverse reward modeling without additional training.


\subsection{Vector Representation of Preferences}
In conventional reward modeling approaches, human preferences are transferred into \textit{scalar scores} using the BT models, as illustrated in Eq. \eqref{eq:bt_loss}. In practice, modern reward models are typically fine-tuned from pretrained or instruction-fine-tuned language models, with a reward head that maps hidden states to a scalar reward prediction. 

In this paper, we consider a \textbf{vector representation} of human preference, similar to recent theoretical work \cite{sun2024rethinking,zhang2024general,shen2025reviving} and empirical studies \cite{sun2023query,sun2025reusing}. We denote $\phi(x, y) \in \mathbb{R}^d$ as a $d$-dimensional feature extractor (e.g., the output from the penultimate layer), and apply a final linear layer $w$ in the reward model, where the reward is computed as $r_{\theta}(x, y) = w^\top \phi(x, y)$.
Under this formulation, the BT objective can be rewritten as:  
\begin{equation*}
\begin{aligned}
  &\max_w  \mathbb{E}_i \left[ \log \sigma \left(w^T \phi(x_i,y_i^c) - w^T \phi(x_i,y_i^r) \right) \right]  \\
  &= \max_w \mathbb{E}_i \left[ \log \sigma \left(w^T (\phi(x_i,y_i^c) - \phi(x_i,y_i^r))\right) \right]
\end{aligned}
\end{equation*}
where \( (x_i,y_i^c), (x_i,y_i^r) \) denote the chosen and rejected samples, respectively. 

This reformulation is particularly interesting because it allows human preference to be captured using a vector \( w \in \mathbb{R}^d \) instead of relying on a large set of model parameters. Consequently, human preference can be interpreted as a direction in the \( d \)-dimensional space. When the feature space difference \( \phi(x_i, y_i^c) - \phi(x_i, y_i^r) \) aligns with such a direction, it means the response pair is aligned with human preference (i.e., \(y_i^c \succ y_i^r\) ); otherwise, it indicates a contradiction. 
Since the distances between preference vectors provide a natural way of comparing preferences, \textit{such a vector-based representation lays a foundation for a more interpretable framework for understanding human preference. }

\subsection{Rethinking Preference Learning with Principal Component Analysis (PCA)}\label{sec:rethink_pca}
The next step is selecting a \textit{good basis} to represent any vector preferences. The basis should be universal, interpretable, and have a strong connection with human preferences. Our choice of the basis is motivated by the following observations.

Formally, we start by defining \( z_i = \phi(x_i,y_i^c) - \phi(x_i,y_i^r) \) and letting it be \( \{z_i\}_{i=1}^{N} \) zero-centered, (i.e., $\mathbb{E}_i [z_i]=0$, which can be ensured via normalization). Without loss of generality, we consider the unified preference vector \( \|w\|_2^2 =1 \). Then the objective of reward modeling becomes
\begin{equation}
\begin{aligned}
      \max_w    \mathbb{E}_i \left[ \log \sigma \left(w^T z_i\right)\right], \quad \text{s.t.} \quad \|w\|_2^2 =1.
\end{aligned}
\end{equation}
Here, \( w \) is optimized to find a direction that best distinguishes the preference data, which bears a resemblance to PCA: both methods seek a meaningful projection of the data onto a dimension. To further explore this connection, consider the following regularized objective ($\lambda > 0$):
\begin{equation}
    J(w) = \mathbb{E}_i  \left[ \log \sigma \left(w^T z_i\right)  \right] - \lambda \|w\|_2^2.
\end{equation}
Taking the gradient, we obtain:
\begin{equation}
    \nabla_w  J(w) = \mathbb{E}_i \left[ (1-\sigma(w^T z_i)) z_i \right] - 2\lambda w.
\end{equation}
For small \( w^T z_i \), we have \( 1-\sigma(x) \approx \frac{1}{2} + c x \) according to Taylor expansion, where \( c \) is a constant. Therefore, we have
\begin{equation}
\begin{aligned}
    \nabla_w  J(w) &\approx c \mathbb{E}_i \left[(z_i z_i^T) w \right] - 2\lambda w \\
    &= c \Sigma w - 2\lambda w,
\end{aligned}
\end{equation}
where \( \Sigma = \mathbb{E}_i \left[ z_i z_i^T \right] \) is the covariance matrix of \( \{z_i\} \). Setting the gradient to zero gives:
\begin{equation}
    c \Sigma w = 2\lambda w \quad \Rightarrow \quad \Sigma w = \frac{2\lambda}{c} w.
\end{equation}
\noindent\textbf{Discussion.} This equation suggests that under certain conditions, the learned preference direction \( w \) aligns with an eigenvector of the covariance matrix \( \Sigma \). While this does not imply a direct equivalence between preference learning and PCA, it highlights an interesting connection: both methods extract a principal direction from the data. Unlike PCA, which maximizes variance in an unsupervised manner, preference learning optimizes a supervised ranking objective, making the relationship approximate rather than exact.

Additionally, while PCA eigenvectors are direction-agnostic—meaning their signs are arbitrary—human preference is inherently directional. As a result, when deriving a preference vector \( w \) from PCA, both \( w \) and \( -w \) need to be considered to ensure an accurate representation.

\subsection{From Scalar Reward to Diverse Rewards}
Our analysis above suggests a connection between the eigenvectors of the covariance matrix and human preferences. A key observation is that the covariance matrix has \( d \) eigenvectors—for example, \( d = 2048 \) for gemma-2B~\cite{team2024gemma} and \( d = 4096 \) for Llama-3.1-8B~\cite{grattafiori2024llama3herdmodels}. This means we can extract a large number of meaningful preference vectors \( w \) from PCA applied to the embedding dataset \( \{z_i\}_{i=1}^{N} \).  

These eigenvectors form an orthonormal basis in the \( d \)-dimensional space, meaning they are mutually orthogonal and span the entire space. Mathematically, the eigendecomposition of the covariance matrix is given by:  
\(
\Sigma = W \Lambda W^T
\)
where \( W = [w_1, w_2, \dots, w_d] \) is an orthonormal matrix whose columns are the eigenvectors of \( \Sigma \), \( \Lambda = \text{diag}(\lambda_1, \lambda_2, \dots, \lambda_d) \) is a diagonal matrix of eigenvalues.  
This ensures that \( w_1, w_2, \dots, w_d \) represent diverse preference directions. Any human preference can then be expressed as a linear combination of these basis vectors:  
\[
w' = \sum_{j=1}^d k_j w_j,
\]
where \( k_1, \ldots, k_d \) are weight parameters. This formulation enables a flexible and expressive human preference representation as a combination of these decomposed rewards.

\subsection{Decomposed Reward Models (DRMs)} \label{sec:drm}
As illustrated in Figure \ref{fig: Framework.}, we propose Decomposed Reward Models (DRMs) by applying PCA to a human preference dataset $\mathcal{D} = \{(x_i, y_i^c, y_i^r)\}_{i=1}^{N}$ using an embedding extractor $\phi(x,y)$. $\phi$ can be any language models (pretrained or instruction-tuned) or reward models that produce hidden states of dimension \( d \). DRMs consist of two main steps:  
(1) \emph{Embedding Extraction:} We run inference over the preference dataset \( \mathcal{D} \) using \( \phi \) to obtain a dataset of embedding differences:  
   \[
   \mathcal{D}_e = \{z_i\}_{i=1}^{N}, \quad \text{where } z_i = \phi(x_i,y_i^c) - \phi(x_i,y_i^r).
   \]
(2) \emph{Principal Decomposition:} Given the dataset \( \mathcal{D}_e \) of shape \( N \times d \), we perform PCA to obtain a set of basis vectors \( W = [w_1, w_2, \dots, w_d] \). Each eigenvector \( w_j \) captures a distinct directional human preference and can be combined with the feature extractor \( \phi(x,y) \) to construct a reward model, leading to a total of $d$ rewards.

To fully utilize DRMs, we aim to represent any human preference as a linear combination of these decomposed rewards. Various methods can achieve this, such as mixture-of-experts \cite{quan2024dmoerm,wang2024interpretable} and test-time alignment \cite{lee2024test}. Since both approaches are fundamentally similar, and test-time alignment can better leverage our method without additional fine-tuning, we adopt it as our implementation. However, \emph{DRMs are compatible with any downstream method that requires multiple diverse reward heads.}

We implement test-time alignment based on HyRe \cite{lee2024test}, which adapts to diverse human preferences by optimizing reward weights given a small adaptation set \( \mathcal{D}_{\rm adapt} \). This is practical in real-world scenarios where new users' data helps refine the strategy to better align with individual preferences. The weights \( k_{m} \) are computed based on the loss function over the adaptation set:  
\begin{equation}
    k_{m} = \frac{\exp{(-\mathcal{L}(w_{m}, \mathcal{D_{\rm adapt}}))}}{ \sum_j\exp{(-\mathcal{L}(w_j, \mathcal{D_{\rm adapt}}))}}
\end{equation}
where
\begin{equation}
    \begin{aligned}
        \mathcal{L}(w_{m}, \mathcal{D_{\rm adapt}})= - \mathbb{E}_{(x_i,y_i^c, y_i^r)\sim \mathcal{D}_{\rm adapt}} \\ \left[  \log \sigma \left(w_{m}^T (\phi(x_i,y_i^c) - \phi(x_i,y_i^r))\right) \right].
    \end{aligned}
\end{equation}
Here, we also normalize the features based on $\mathcal{D}_{\rm adapt}$ to prevent the loss $\mathcal{L}$ from being dominated by samples with large-scale features.
This formulation ensures that weights \( k_m \) dynamically adjust based on the dataset, assigning higher importance to directions \( w_m \) with lower loss and reducing the influence of those with higher loss. As a result, we can obtain a soft, weighted sum of DRMs that better aligns with the desired human preference.

\paragraph{Advantages of DRMs.}
(1) \emph{Simplicity:} Our proposed DRMs offer a simple yet effective approach to reward modeling without requiring additional training. (2) \emph{Diversity:} Unlike traditional scalar reward models that struggle to represent heterogeneous human preferences, DRMs leverage a diverse set of basis vectors to capture a wide range of preferences. (3) \emph{Adaptivity:} By decomposing human preference data through PCA, DRMs naturally extract meaningful directional preferences, allowing them to adapt flexibly to different user needs. This adaptability can be leveraged by test-time alignment, which dynamically adjusts reward weights to better suit specific preferences. As a result, DRMs provide a scalable and interpretable solution for modeling complex and diverse human preferences.

\begin{table*}[h!]
    \centering
    \renewcommand{\arraystretch}{1.2}
    \setlength{\tabcolsep}{4pt}
    \resizebox{0.95\linewidth}{!}{%
    \begin{tabular}{p{2.2cm}|l|c|c|c|c|c|c}
    \toprule
    \multirow{2}{*}{\textbf{Benchmark}} & \multirow{2}{*}{\textbf{Attribute}} & \multicolumn{3}{c|}{\textbf{Gemma-2B-RM}} & \multicolumn{3}{c}{\textbf{Llama3-8B-RM}} \\
    \cline{3-8}
    & & Single Head & Max Value & Max Head & Single Head & Max Value & Max Head \\
    \midrule
    \multirow{5}{*}{RewardBench} 
    & Overall & 0.733 & \textbf{0.735} & head\_0 & 0.862 & \textbf{0.869} & head\_0 \\
    & Chat & 0.944 & \textbf{0.950} & head\_0 & 0.983 & \textbf{0.986} & head\_0 \\
    & Chat Hard & 0.467 & \textbf{0.660} & head\_3 & 0.684 & \textbf{0.695} & head\_3 \\
    & Safety & \textbf{0.759} & 0.745 & head\_0, head\_8 & 0.868 & \textbf{0.886} & head\_0 \\
    & Reasoning & 0.759 & \textbf{0.821} & head\_32 & 0.912 & \textbf{0.923} & head\_0 \\
    \midrule
    \multirow{6}{*}{RPR} 
    & Overall & 0.714 & \textbf{0.735} & head\_0 & \textbf{0.853} & 0.839 & head\_0 \\
    & User-Friendliness & 0.506 & \textbf{0.798} & head\_9, head\_26 & 0.719 & \textbf{0.899} & head\_10 \\
    & Narrative \& Storytelling & 0.662 & \textbf{0.825} & head\_12 & 0.838 & \textbf{0.912} & head\_5 \\
    & Linguistic Creativity & 0.817 & \textbf{0.885} & head\_12 & 0.875 & \textbf{0.981} & head\_37 \\
    & Scientific Rigor & \textbf{0.881} & \textbf{0.881} & head\_34 & 0.940 & \textbf{0.964} & head\_0 \\
    & Humor \& Entertainment & 0.690 & \textbf{0.964} & head\_9 & 0.893 & \textbf{0.952} & head\_37, head\_74 \\
    \bottomrule
    \end{tabular}
    }
    \caption{Performance of top 100 decomposed reward heads. ``Single Head'' is the trained single-head baseline. ``Max Value'' refers to the highest score achieved for each attribute, while ``Max Head''  indicates which specific head attains this maximum score. ``Overall'' represents the average accuracy of a single head across all attributes.}
    \label{single_head_analysis}
\end{table*}

\section{Experiment}
In this section, we conduct extensive experiments to evaluate the effectiveness of DRMs, focusing on the \emph{diversity and interpretability} of the decomposed heads, as well as their \emph{adaptivity} to downstream human preferences.

\subsection{Experimental Setup}

\textbf{Dataset.} We choose the \textit{mixture2\_and\_safe\_pku} dataset\footnote{\url{https://huggingface.co/datasets/weqweasdas/preference_dataset_mixture2_and_safe\_pku}}~\cite{dong2024rlhf}, a collection of 550k pairwise preference samples. This dataset combines diverse data sources, including human-labeled preferences from HH-RLHF~\cite{bai2022training} and GPT-labeled preferences from UltraFeedback~\cite{cui2023ultrafeedback}, making it well-suited for studying diverse preference decomposition. 
To evaluate the effectiveness of the decomposed reward heads, we test models on two unseen benchmarks with multiple attributes: (1) RewardBench~\cite{lambert2024rewardbench}, a dataset designed to evaluate reward models across various dimensions, including chat quality, safety, and reasoning. (2) Reasonable Preference Reversal (RPR) test set~\cite{pitis2024improving} focuses on personalized context-aware preference evaluation. From RPR, we sample five fine-grained categories (i.e., User-Friendliness, Narrative Quality, Linguistic Creativity, Scientific Rigor, and Humor), each with over 80 annotated samples. Other categories were excluded due to insufficient data for reliable evaluation.

\noindent\textbf{Base Model.} 
For experiments on decomposed reward heads and test-time adaptation, we use two open-source reward models, \href{https://huggingface.co/Ray2333/Gemma-2B-rewardmodel}{Gemma-2B-RM} and \href{https://huggingface.co/Ray2333/GRM-llama3-8B-distill}{Llama3-8B-RM}~\cite{yang2024regularizing}, along with an instruction-tuned language model, \href{https://huggingface.co/google/gemma-2-9b-it}{gemma-2-9b-it} \cite{team2024gemma}, as our base models.  
To analyze their performance, we keep the backbone fixed as our feature extractors while generating multiple new reward heads for them.

\noindent\textbf{Baselines.} We compare the performance of DRMs against several baselines. (1) \textit{Single-Head RM} fine-tunes a single reward head on top of a fixed backbone using the same dataset as DRMs. (2) \textit{Share-Base RM}~\cite{lee2024test} is a training-based method that employs multiple learnable reward heads while incorporating a frozen prior network\cite{osband2023epistemic} to maintain diversity, with the final output derived from their combination. (3) \textit{Random Head} uses multiple reward heads with randomly initialized weights to capture diverse but largely unstructured preferences. Specifically, we experiment with both uniform and Gaussian initialization, followed by L2 normalization to ensure consistency with DRM vectors. 


\begin{table*}[h]
    \centering
    \renewcommand{\arraystretch}{1.2}
    \setlength{\tabcolsep}{6pt}  
    \resizebox{0.95\linewidth}{!}{  
    \begin{tabular}{p{2.45cm} l c c c c c c}  
        \toprule
        \textbf{Base Model} & \textbf{Method} & \shortstack{\textbf{User} \\ \textbf{Friendliness}} & \shortstack{\textbf{Narrative} \\ \textbf{\& Storytelling}} & \shortstack{\textbf{Linguistic} \\ \textbf{Creativity}} & \shortstack{\textbf{Scientific} \\ \textbf{Rigor}} & \shortstack{\textbf{Humor} \\ \textbf{\& Entertainment}} & \textbf{Overall} \\

        \midrule
        \multirow{4}{*}{\textbf{Gemma-2B-RM}} 
        & Single Head
            & 0.506 & 0.662 & 0.817 & 0.881 & 0.690 & 0.714 \\
        & Shared-Base  
            & 0.517(0.000) & 0.688(0.000) & 0.817(0.000) & 0.881(0.000) & 0.690(0.000) & 0.721(0.000) \\
        & Random (Uniform)  
            & 0.713(0.062) & 0.782(0.068) & 0.920(0.045) & 0.907(0.043) & 0.907(0.026) & 0.848(0.024) \\
        & Random (Gaussian)  
            & 0.582(0.039) & 0.760(0.060) & 0.823(0.063) & 0.873(0.025) & 0.817(0.053) & 0.771(0.022) \\
        \rowcolor{gray!15}
        & \textbf{DRM (Ours)}  
            & \textbf{0.789(0.062)} & \textbf{0.871(0.033)} & \textbf{0.953(0.034)} & \textbf{0.907(0.019)} & \textbf{0.975(0.017)} & \textbf{0.900(0.017)} \\
        \midrule
        \multirow{4}{*}{\textbf{Llama3-8B-RM}}  
        & Single Head  
            & 0.685 & 0.825 & 0.846 & 0.964 & 0.905 & 0.844 \\ 
        & Shared-Base  
            & 0.674(0.000) & 0.825(0.000) & 0.827(0.000) & 0.964(0.000) & 0.881(0.000) & 0.832(0.000) \\
        & Random (Uniform)  
            & 0.616(0.104) & 0.860(0.037) & 0.798(0.103) & 0.958(0.007) & 0.906(0.031) & 0.823(0.041) \\
        & Random (Gaussian)  
            & 0.730(0.100) & 0.892(0.027) & 0.887(0.055) & 0.956(0.008) & 0.919(0.032) & 0.875(0.028) \\
        \rowcolor{gray!15}
        & \textbf{DRM (Ours)}  
            & \textbf{0.812(0.063)} & \textbf{0.946(0.029)} & \textbf{0.945(0.015)} & \textbf{0.969(0.010)} & \textbf{0.991(0.011)} & \textbf{0.931(0.016)} \\
        \bottomrule
    \end{tabular}
    }
    \caption{Evaluation Results on RPR ($n=5$). We compare DRMs with trained baselines (``Single Head'' and ``Shared-Base''), and randomly generated multi-head baselines (``Random''). Except for single-head baseline, other methods use HyRe for test-time adaptation. Standard deviation over 20 sampled adaptation sets are reported.}\label{tab:RPR_res}
\end{table*}

\begin{table*}[h]
    \centering
    \small
    \renewcommand{\arraystretch}{1.2}
    \setlength{\tabcolsep}{6pt}  
    \resizebox{0.95\textwidth}{!}{
    \begin{tabular}{p{2.2cm} l c c c c c}  
        \toprule
        \textbf{Base Model} & \textbf{Method} & \textbf{Chat} & \textbf{Chat Hard} & \textbf{Safety} & \textbf{Reasoning} & \textbf{Overall} \\
        \midrule
        \multirow{6}{*}{\textbf{Gemma-2B-RM}} 
        & Single Head 
            & 0.944 & 0.467 & 0.759 & 0.759 & 0.733 \\
        & Shared-Base 
            & 0.947(0.000) & 0.476(0.000) & 0.765(0.000) & 0.774(0.000) & 0.740(0.000) \\
        & Random (Uniform)  
            & 0.940(0.005) & 0.567(0.029) & \textbf{0.800(0.010)} & 0.843(0.019) & 0.787(0.009) \\
        & Random (Gaussian)  
            & 0.951(0.005) & 0.573(0.033) & 0.781(0.015) & 0.839(0.021) & 0.786(0.008) \\
        \rowcolor{gray!15}
        & DRMs(Ours) 
            & \textbf{0.953(0.003)} & \textbf{0.650(0.028)} & 0.783(0.030) & \textbf{0.872(0.025)} & \textbf{0.814(0.013)} \\
        \midrule
        \multirow{4}{*}{\textbf{Llama3-8B-RM}}  
        & Single Head  & \textbf{0.989} & 0.684 & 0.891 & 0.920 & 0.871 \\ 
        & Shared-Base   & 0.986(0.000) & 0.684(0.000) & 0.895(0.000) & 0.927(0.000) & 0.873(0.000) \\ 
        & Random (Uniform)  
            & 0.985(0.003) & 0.623(0.089) & \textbf{0.903(0.010)} & 0.915(0.014) & 0.857(0.023) \\
        & Random (Gaussian)  
            & 0.982(0.004) & 0.663(0.096) & 0.889(0.009) & \textbf{0.936(0.011)} & 0.868(0.024) \\
        \rowcolor{gray!15}
        & DRMs(Ours) & 0.986(0.002) & \textbf{0.755(0.032)} & 0.885(0.036) & 0.914(0.036) & \textbf{0.885(0.012)} \\
        \bottomrule
    \end{tabular}}
    \caption{Evaluation Results ($n=15$), comparing different methods across two base models. }\label{tab:eval_rmbench}
\end{table*}

\begin{table}[h]
    \centering 
    \renewcommand{\arraystretch}{1.2} 
    \resizebox{\linewidth}{!}{%
    \begin{tabular}{l|l|c|c|c}
    \toprule
  \multirow{2}{*}{  \textbf{Benchmark}} & \multirow{2}{*}{\textbf{Attributes}} & \textbf{Single} & \textbf{Random}  & \textbf{DRMs}\\
    &   & \textbf{Head}  & \textbf{(Uniform)}  & \textbf{(Ours)}  \\
    \midrule
     \multirow{5}{*}{\cellcolor{white}\textbf{RewardBench}} 
        & Overall    & 0.759 & 0.770 & \textbf{0.830} \\
        & Chat       & 0.905 & 0.897 &\textbf{ 0.920 }\\
        & Chat Hard  & 0.621 & 0.600 & \textbf{0.692} \\
        & Safety     & 0.699 & 0.753 & \textbf{0.786} \\
        & Reasoning  & 0.813 & 0.832 & \textbf{0.920 } \\
    \midrule 
    \multirow{6}{*}{\cellcolor{white}\textbf{RPR}} 
        & Overall                     & 0.746 & 0.630 & \textbf{0.796} \\
        & User-Friendliness          & 0.640 & 0.555 & \textbf{0.657} \\
        & Narrative \& Storytelling & 0.713 & 0.610 & \textbf{0.763} \\
        & Linguistic Creativity      & 0.808 & 0.595 & \textbf{0.843} \\
        & Scientific Rigor           & 0.762 & 0.661 & \textbf{0.806} \\
        & Humor \& Entertainment    & 0.798 & 0.744 & \textbf{0.905} \\
    \bottomrule
    \end{tabular}
    }
    \caption{Performance of DRMs on \textbf{instruction-tuned model} 
    Gemma-2-9b-it. Single head baseline is trained with the same dataset used for DRMs. Aligned with the previous setting, we use $n=15$ and $n=5$ for RewardBench and RPR respectively.}\label{tab:gemma-2-9b-it_exp}
    \vspace{-20pt}
\end{table}

\subsection{What information is Captured by DRMs?}
\label{single_head}
We aim to better understand the decomposed reward heads in DRMs. To achieve this, we evaluate the performance of the top 100 reward vectors, ranked by eigenvalue, on both RewardBench and RPR’s fine-grained subsets.
Table~\ref{single_head_analysis} reports the scores of a trained single-head baseline. The ``Max Value'' column shows the highest score achieved for each attribute, while the ``Max Head'' column indicates which reward head achieves this score. We also compare the results with the single-head baseline. The results reveal the following findings:

\noindent (1) \textbf{Diversity and Interpretability:} DRMs effectively capture diverse human preferences, with different reward heads excelling at different attributes. For instance, in Gemma-2B-RM, head\_9 performs best on ``User-Friendliness'' (accuracy: 0.798) and ``Humor and Entertainment'' (0.964), while head\_12 excels in ``Narrative and Storytelling'' (0.825) and ``Linguistic Creativity'' (0.885). In contrast, the single-head RM fails to capture this diversity, yielding suboptimal performance on attributes such as ``User-Friendliness'' (0.506) and ``Humor and Entertainment'' (0.690) for Gemma-2B-RM. These results indicate that DRMs not only capture a broader range of human preferences but also provide interpretable representations that align well with certain preference attributes.

\noindent (2) \textbf{The first head is the most informative:} An interesting observation is that the head head\_0 consistently achieves the highest overall accuracy for both models. This aligns with expectations, as head\_0 corresponds to the eigenvector with the largest variance, i.e., the most informative direction. Furthermore, among the top 100 heads, most of the high-performing heads appear before index 40, which aligns with PCA’s property that the explained variance decreases as the head index increases. This finding further supports our argument that PCA can approximate preference learning.

In summary, our results show that a single reward head is insufficient to represent the full spectrum of human preferences. Instead, DRMs provide high-quality and interpretable estimations of diverse human preferences, supporting our analysis in Section \ref{sec:rethink_pca}.

\vspace{-8pt}
\subsection{Test-time Preference Adaptation}
\label{TTA}
\vspace{-2pt}
A natural application of DRMs is test-time adaptation—deriving linear combinations to match new user preferences. Following the adaptation method in Section \ref{sec:drm}, we use a small subset of test data for each attribute (e.g., $n=15$ for RewardBench and $n=5$ for RPR), which corresponds to less than $4\%$ of the available data per attribute in RewardBench and less than $6\%$ for RPR. We compare DRMs against several baselines, including the single-head and ensemble-head baselines trained on the same dataset, and two random-head baselines that sample random heads. To ensure efficiency, we limit all models, including DRMs, to using only 100 heads. The results for the two benchmarks, shown in Table~\ref{tab:RPR_res} and Table~\ref{tab:eval_rmbench}, include the standard deviation over 20 repetitions of sampled adaptation sets.

The results demonstrate that DRMs achieve the best overall performance across different base models and test sets. The improvement is particularly significant for Gemma-2B-RM, where DRMs improve the single-head baseline from 0.733 to 0.814 on RewardBench and from 0.714 to 0.90 on RPR. Interestingly, the ``Shared-Base'' ensemble baseline does not outperform the single-head baseline, suggesting that it lacks diversity in its learned reward heads. In contrast, the random-head baseline offers some improvement over the single-head baseline but remains inferior to DRMs. This confirms that DRMs provide a diverse and well-structured set of basis vectors that enable efficient test-time adaptation to user preferences.

Another important question is whether a language model can serve as a feature extractor instead of a reward model. To explore this, we conduct experiments using Gemma-2-9B-it as the feature extractor. The results, presented in Table~\ref{tab:gemma-2-9b-it_exp}, show that DRMs can successfully exploit a language model for this purpose, outperforming the single-head trained baseline by 7.8\% on RewardBench and 26\% on RPR. Furthermore, comparing results across models reveals that while DRMs perform well with language models as feature extractors, reward models remain a more effective choice.

\begin{figure}[t]
\vspace{-5pt}
    \centering
    \includegraphics[trim={0 0cm 0 2cm}, clip, width=1\linewidth]{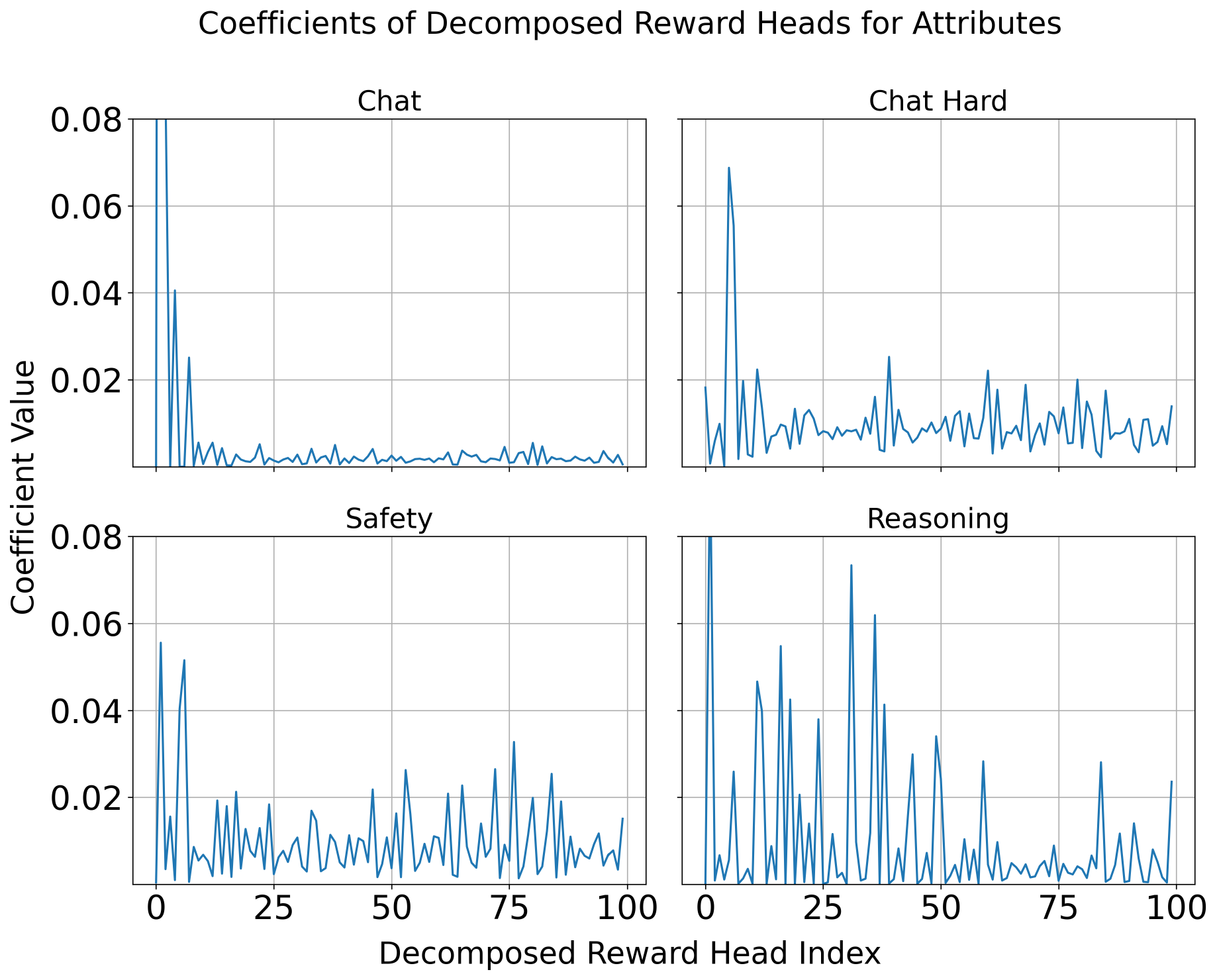}
    \vspace{-10pt}
    \caption{Weight distributions of the top 100 decomposed reward heads on RewardBench for DRMs using Gemma-2B-RM as the backbone.}
    \label{fig:weight_distribution}
    \vspace{-10pt}
\end{figure}

\begin{figure}[t]
\vspace{-10pt}
    \centering
    \includegraphics[trim={2.5cm 0.7cm 2.2cm 0}, clip, width=1\linewidth]{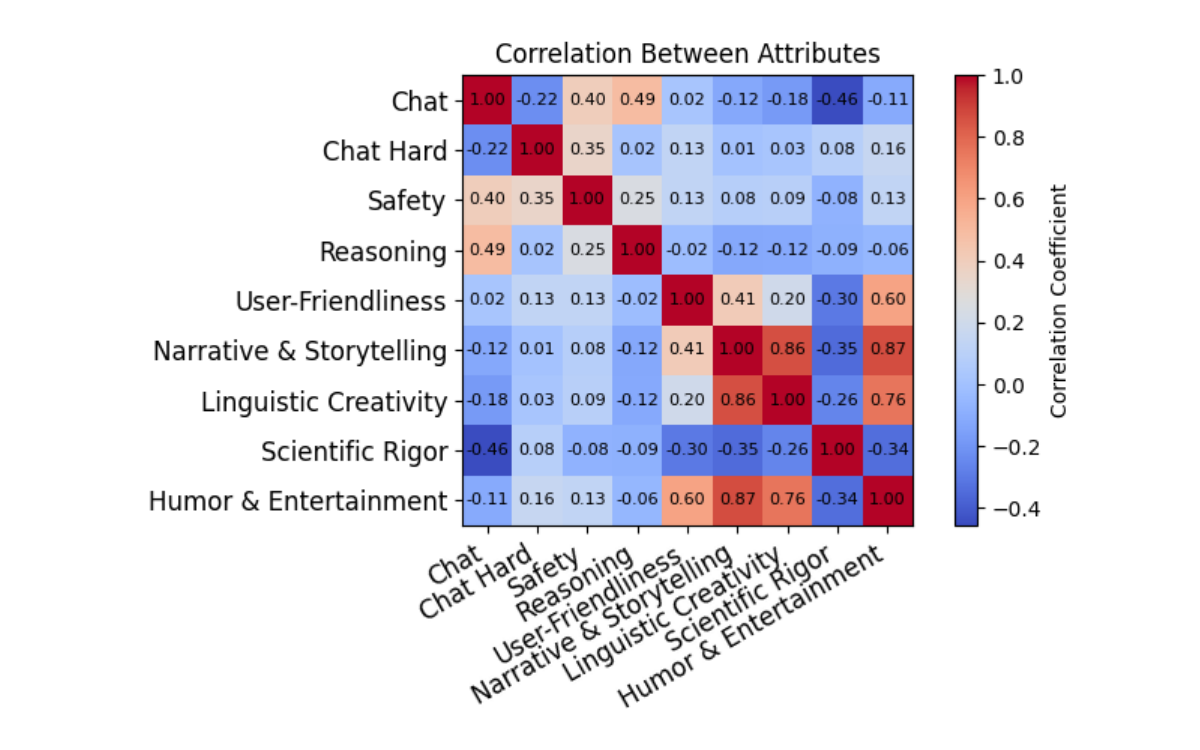}
    \vspace{-20pt}
    \caption{Correlation among attributes' weight vectors for DRMs. The feature extractor is Gemma-2B-RM.}
    \label{fig:Correlation}
    \vspace{-5pt}
\end{figure}

\subsection{Quantitative Attribute Explainability}\label{sec:attribute_interp}
Beyond its adaptability, DRMs offer a significant advantage in interpretability, helping not only to understand human preferences but also to analyze the multiple attributes present in current test sets. Specifically, for each attribute subset, we can obtain the weight parameters $\textbf{k} = [k_1, \ldots,k_d]$ corresponding to each basis vector. Figure~\ref{fig:weight_distribution} visualizes some of these weight vectors in RewardBench, revealing distinct patterns. The \emph{Chat} subset primarily relies on the first few basis vectors, which capture the most data variance. In contrast, the \emph{Chat Hard} and \emph{Safety} subsets exhibit a more uniform weight distribution, while the \emph{Reasoning} subset depends more heavily on basis vectors indexed between 1 and 50. These variations highlight the differing preference requirements across subsets. 

Using the weight vectors $\textbf{k}$ for all attributes, we compute their Pearson correlation, providing a quantitative explanation of the dataset's attributes. The resulting correlation matrix, shown in Figure~\ref{fig:Correlation}, reveals meaningful relationships between attributes. For instance, ``\emph{Narrative \& Storytelling}'' strongly correlates with ``\emph{Humor \& Entertainment}'' and ``\emph{Linguistic Creativity}'' (with a correlation of approximately 0.87), which aligns with the idea that humor and creativity enhance storytelling. This correlation suggests that some attributes may be redundant. 
On the other hand, ``\emph{Scientific Rigor}'' negatively correlates with several attributes, including ``\emph{Chat}'' and ``\emph{Narrative \& Storytelling}'' (\(\text{Pearson } r = -0.46\) and \(-0.35\), respectively), suggesting that scientific rigor may conflict with other human preferences. Additionally, many attributes show weak or negligible correlations (with absolute Pearson \(r < 0.1\)).
Overall, DRMs provide a structured framework for quantitatively explaining attribute relationships, offering deeper insights into benchmark design and multi-attribute evaluation.

\subsection{Ablation Study}\label{sec:ablation}
\begin{figure}[t]
    \centering
    \begin{subfigure}[b]{0.49\linewidth}
        \centering
        \includegraphics[width=\linewidth]{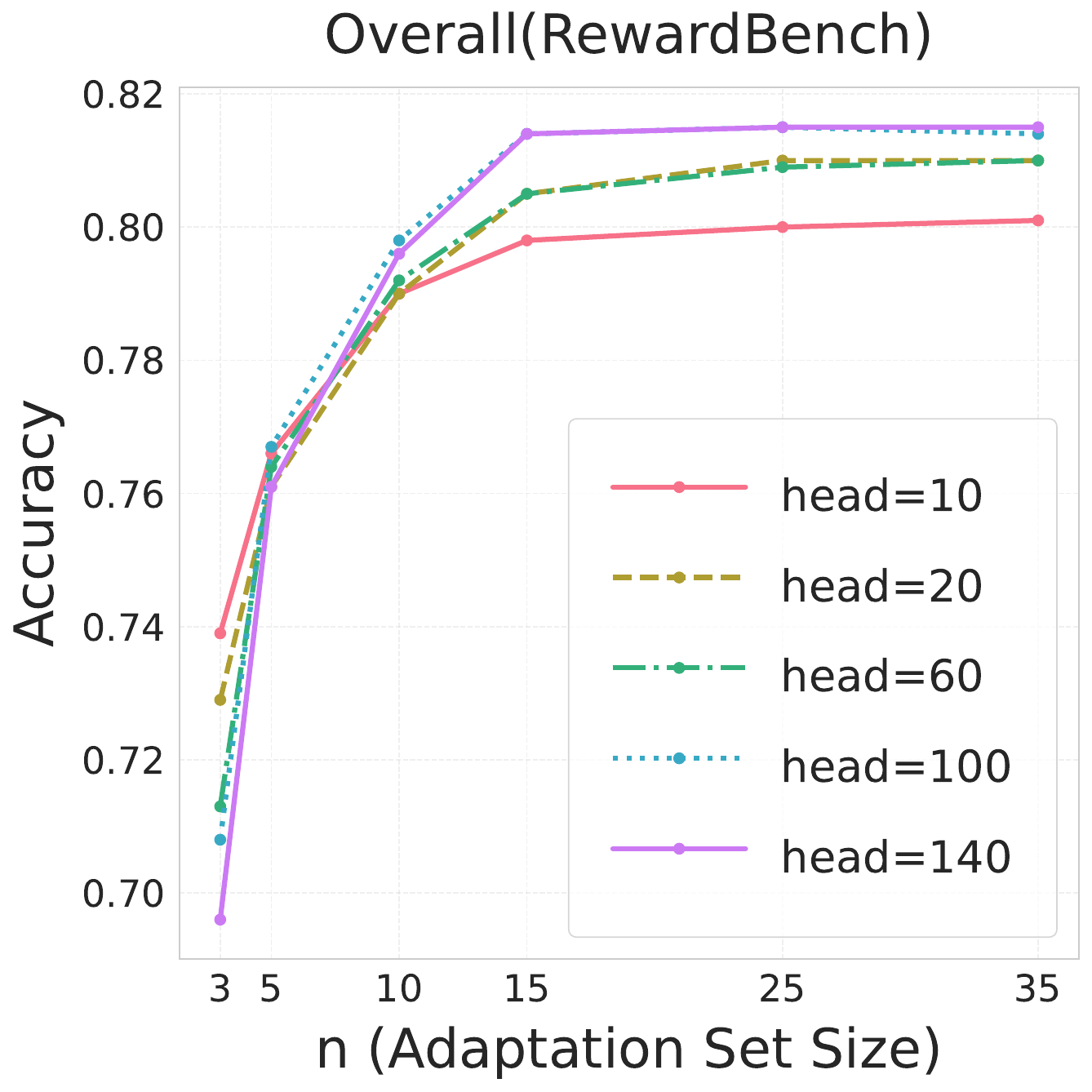}
        \label{fig:fig1}
    \end{subfigure}
    \begin{subfigure}[b]{0.49\linewidth}
        \centering
        \includegraphics[width=\linewidth]{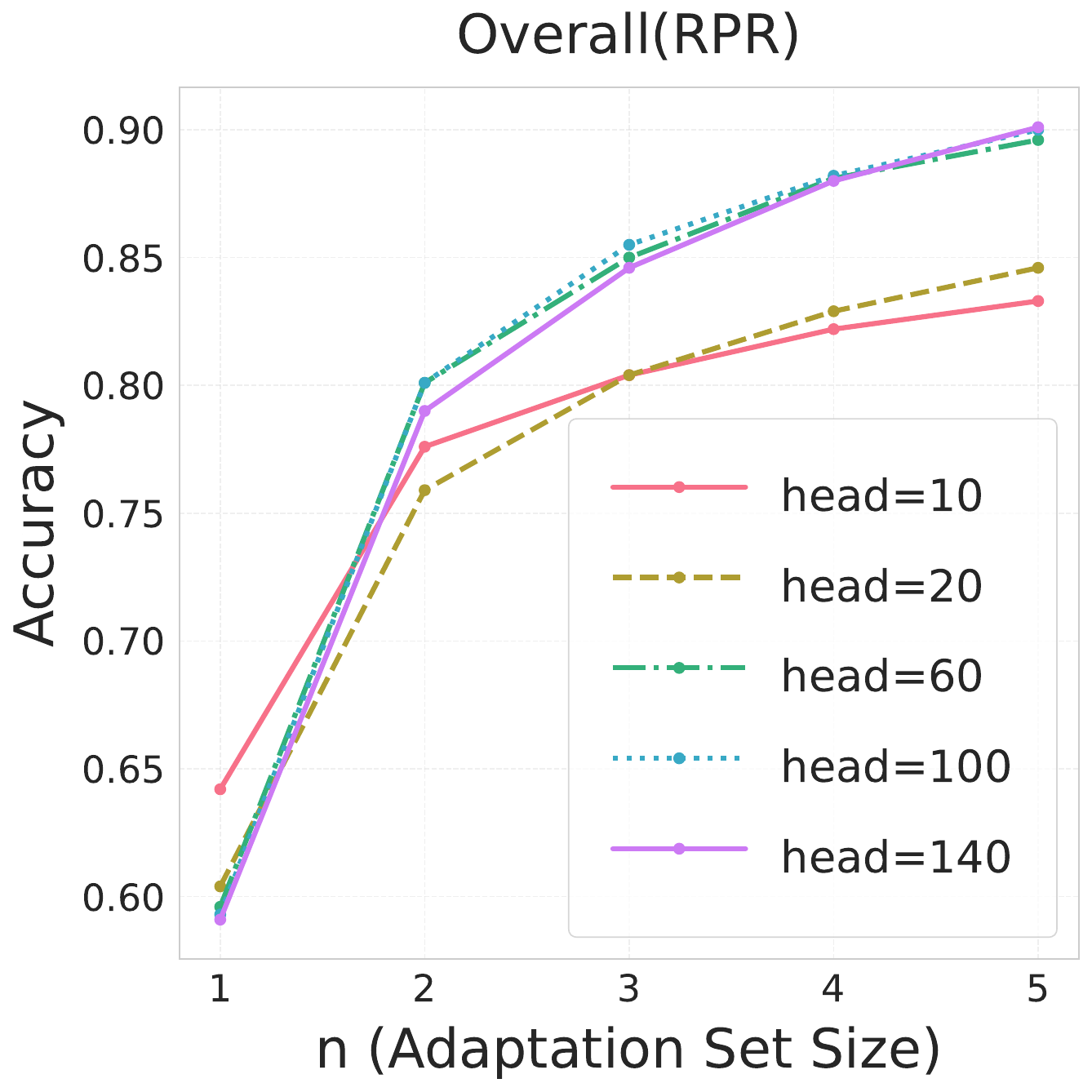}
        \label{fig:fig2}
    \end{subfigure}
    \vspace{-25pt}
    \caption{Ablations on the adaptation set size and number of reward heads for test-time adaptation based on Gemma-2B-RM. }
    \label{fig:ablation}
    \vspace{-5pt}
\end{figure}

We analyze two key factors affecting test-time adaptation: adaptation set size and the number of DRM heads used. Using Gemma-2B-RM as the feature extractor, we present results on RPR and RewardBench in Figure \ref{fig:ablation}.  
Our findings show that performance improves with a larger adaptation set, converging on RewardBench at \( n \geq 15 \). Similarly, increasing the number of heads enhances performance but saturates beyond 100, likely because the most meaningful PCA directions lie within the first 100 heads.  
When the adaptation set is small (e.g., \( n = 3 \)), performance is unstable, and fewer heads can yield better results. This may be due to difficulty in correctly weighting heads with limited data, whereas using more heads increases the risk of assigning incorrect weights. However, with sufficient data, more heads eventually lead to better performance. These results suggest that a slightly larger adaptation set and a carefully chosen number of heads are key to optimizing performance.



\section{Related work}
\noindent\textbf{The Heterogeneity of Reward Modeling.} 
Reward models \cite{lambert2024rewardbench,liu2024skywork} are typically trained using preference annotations from the Bradley-Terry model~\citep{christiano2017deep,bradley1952rank} or demonstration data~\citep{wulfmeier2024imitating,xiao2024leverage}. However, human preferences are diverse and complex, making it difficult to capture all relevant attributes with a single objective~\citep{yang2024rewards,rame2024rewarded,chakraborty2024maxmin}. To address this, researchers are exploring multi-objective preference learning. This includes collecting datasets that assess multiple attributes~\citep{wu2023fine,wang2023helpsteer,cui2023ultrafeedback,pitis2024improving} and developing multi-head reward models that learn diverse user preferences~\citep{quan2024dmoerm,wang2024interpretable}.


\noindent\textbf{Embedding-based Reward Model.} 
Recent advances in reward modeling have highlighted embedding-based approaches for their efficiency and scalability~\citep{ahmed2024scalable,sun2023query,zhang2024general}. These models are backed by strong theoretical foundations~\citep{sun2024rethinking} and demonstrate high flexibility with competitive or superior performance~\citep{li2024q,tennenholtz2024embedding}. Additionally, they integrate well with established statistical learning tools~\citep{dykstra1960rank,springall1973response,han2020asymptotic} and offer greater transparency through statistical insights~\citep{shen2025reviving,feng2025pilaf}. Unlike prior methods, our approach represents human preferences through the final linear layer, enhancing interpretability and enabling the decomposition of preference components.

\noindent\textbf{Dimensionality Reduction and Embedding Analysis.}
Linear dimensionality reduction techniques like PCA have proven effective in extracting latent dimensions that capture key human preferences for model alignment~\citep{freire2024uncovering}. Beyond alignment, broader studies have explored structuring high-dimensional data into more interpretable embeddings~\cite {huertas2023exploring,kanerva2000random}. Methods such as Q-Probe~\citep{li2024q} and DeepMDP~\citep{gelada2019deepmdp} further enhance model alignment by efficiently exploring embedding spaces. Recent work also shows that improving embedding quality leads to better model performance and generalization~\citep{li2024reward,yang2024regularizing}.

\section{Conclusion}
In this paper, we establish the connection between preference learning and PCA, introducing Decomposed Reward Models (DRMs). DRMs represent diverse human preferences as a set of orthogonal basis vectors using a novel vector-based formulation of preference. This approach enables efficient test-time adaptation to user preferences without requiring additional training, making it both scalable and practical.
Beyond the efficiency, DRMs provide a structured way to understand human preferences. By decomposing complex preferences into interpretable components, they reveal how preferences are formed and interact. We hope this work inspires further research into the fundamentals of human preference learning while promoting more transparent and personalized AI systems.

\section{Limitations}
In this paper, we obtain a large number of decomposed rewards from DRMs. However, due to the large scale (e.g., 2048 or 4096 reward heads), we did not manually examine each head to identify its corresponding preference attribute. Future work could focus on developing automated methods to analyze these rewards, such as recognizing patterns in the first 100 reward heads and assessing whether the last 100 primarily capture noise or meaningful subtleties.  
Additionally, we did not incorporate interdisciplinary study by collaborating with psychology experts to explore human preferences in depth. Future research could benefit from such collaboration to bridge the gap between computational models and cognitive science.

\section{Ethics Statement}
This paper introduces Decomposed Reward Models (DRMs), a step toward improving multi-objective alignment in LLMs. Here, we discuss the potential benefits of our approach while acknowledging the associated risks.  

Our method enhances LLM alignment with diverse human preferences. DRMs are lightweight, flexible, and easily adaptable to new users and evolving preferences. Their efficiency reduces resource demands and broadens accessibility, paving the way for personalized preference learning and scalable LLM alignment. By offering an interpretable framework, DRMs promote greater transparency and customization in human preference modeling.

We carefully follow the license for all datasets and models used in our paper. Human preference datasets often contain biases, reflecting the perspectives and prejudices of their sources. If not properly managed, these biases could propagate through the model, influencing decomposition and principal components, potentially leading to unintended consequences. Mitigating this risk requires careful curation, filtering, and bias reduction before large-scale deployment. Additionally, our method does not inherently control the meaning of each reward head, which could unintentionally capture harmful human preferences. Therefore, thorough evaluation is necessary before deployment to ensure ethical and responsible use.

\section{Acknowledgements}
We thank the anonymous reviewers for their valuable comments. We thank the Chili Lab at Rice for helpful discussions and suggestions throughout this research.

\newpage



\appendix

\section{Appendix}
\label{sec:appendix}

\subsection{Ablations on Llama3-8B-RM}
\begin{figure}[h]
    \centering
    \begin{subfigure}[b]{0.49\linewidth}
        \centering
        \includegraphics[width=\linewidth]{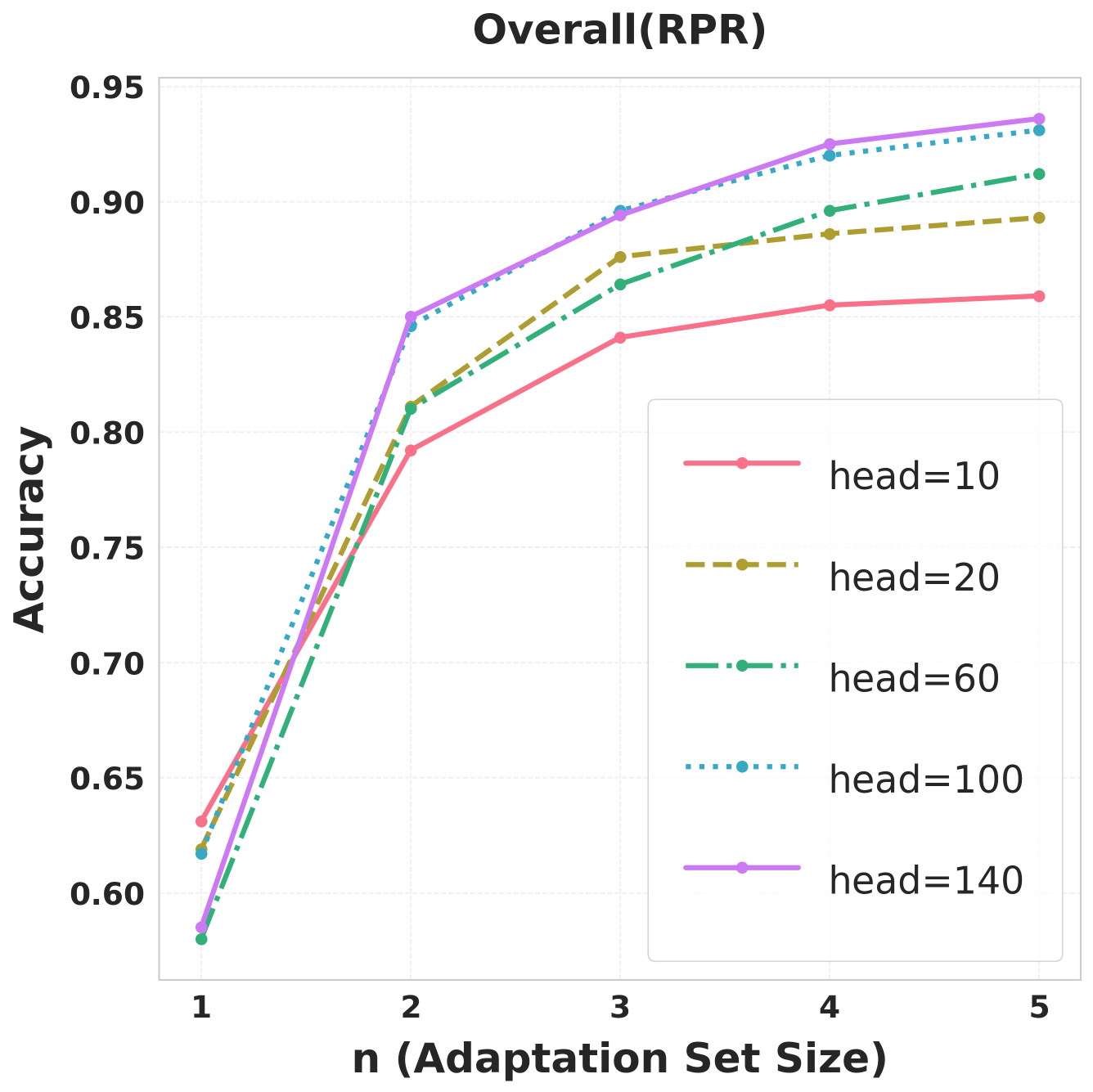}
        \label{fig:fig_2_1}
    \end{subfigure}
    \begin{subfigure}[b]{0.49\linewidth}
        \centering
        \includegraphics[width=\linewidth]{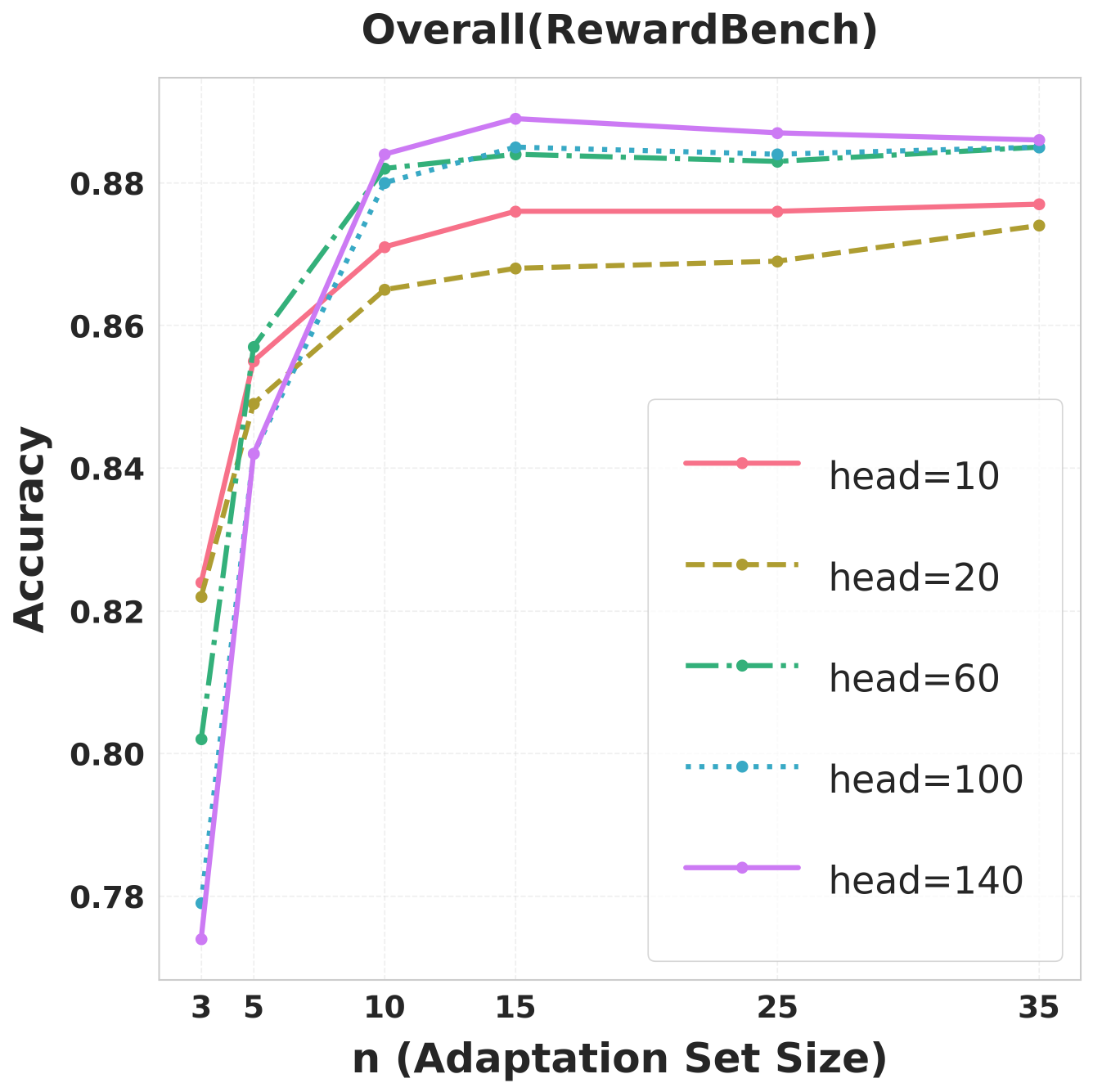}
        \label{fig:fig_2_2}
    \end{subfigure}
    \caption{Ablations on the adaptation set size and number of reward heads for test-time adaptation on Llama3-8B-RM. }
    \label{fig:ablation_8b}
\end{figure}

We add the ablation results on Llama3-8B-RM in Figure \ref{fig:ablation_8b}. The trend is similar to the ablations in our main paper.

\begin{figure}[h]
    \centering    \includegraphics[width=1.0\linewidth, trim=52 40 52 85, clip]{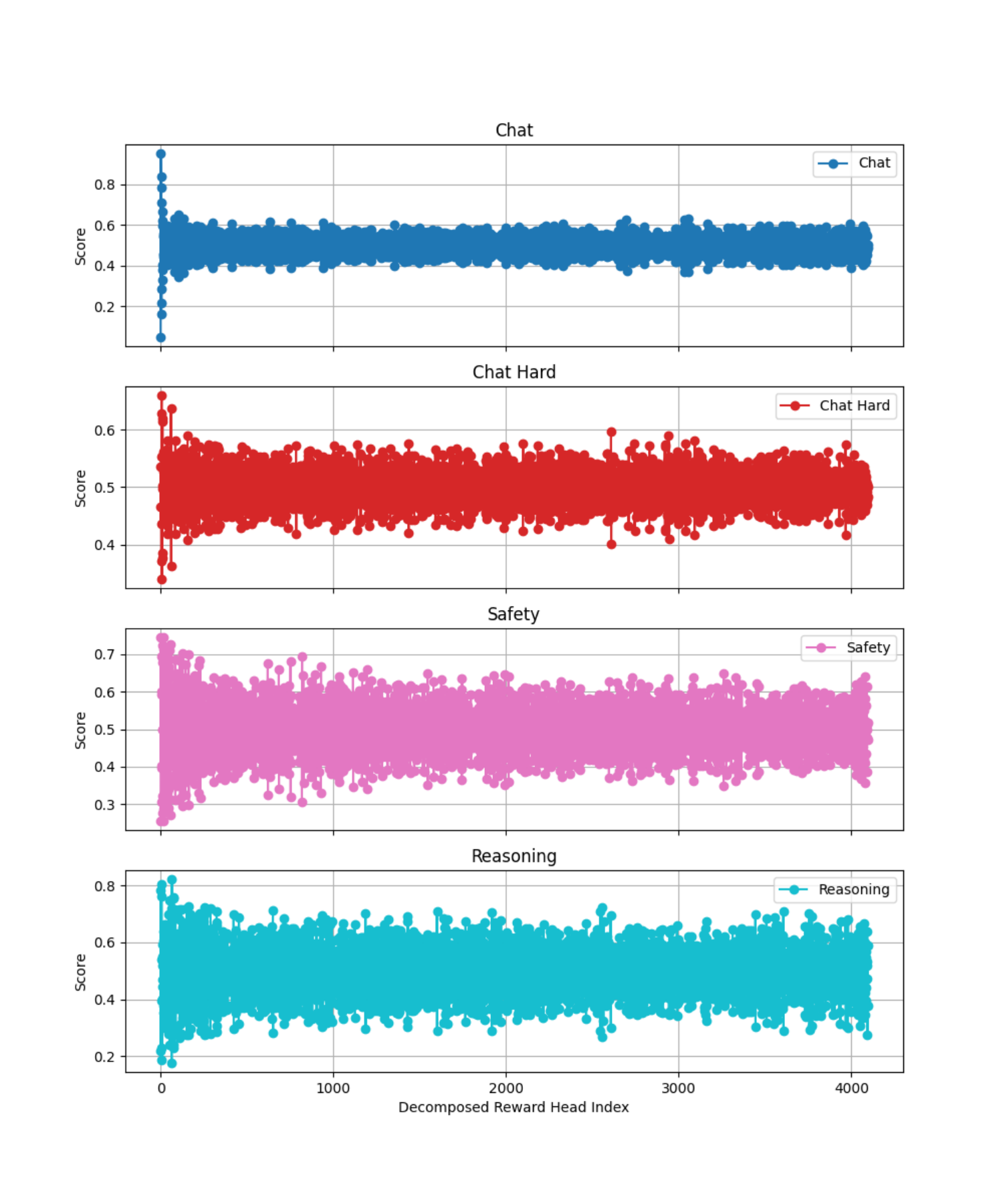}
    \caption{Reward scores of all decomposed reward heads on RewardBench for DRMs using Gemma-2B-RM as the backbone.}
    \label{fig:rewardscores_2b_rb}
\end{figure}
\subsection{Reward Scores on decomposed reward heads}
\begin{figure}[h]
    \centering
    \includegraphics[width=1.0\linewidth]{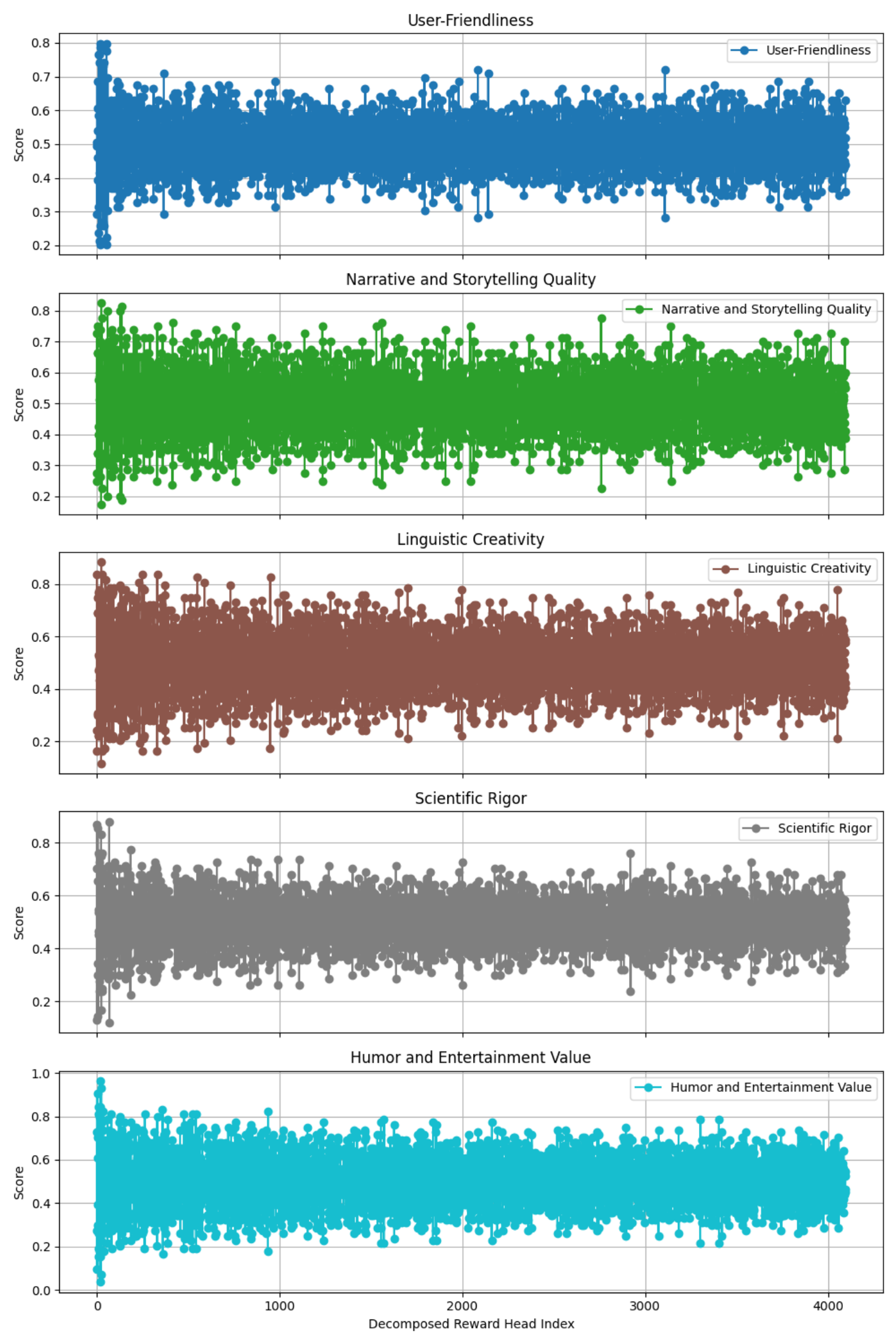}
    \caption{Reward scores of all decomposed reward heads on RPR for DRMs using Gemma-2B-RM as the backbone.}
    \label{fig:rewardscores_2b_rpr}
\end{figure}

As shown in Figure \ref{fig:rewardscores_2b_rb} and \ref{fig:rewardscores_2b_rpr}, we visualize the reward scores of individual decomposed reward heads on both RewardBench and RPR, respectively. With a hidden dimension of 2048 in Gemma-2B-RM, the total number of reward heads is 4096. While most heads' scores fall within a certain range, a few outliers are identified, which will be utilized during test-time preference adaptation for specific tasks.

\subsection{Implementation Details}
For all training-based reward head models, including both the Single Head and Share-Base variants, we train them on the \textit{mixture2\_and\_safe\_pku} dataset\footnote{\url{https://huggingface.co/datasets/weqweasdas/preference_dataset_mixture2_and_safe_pku}}~\cite{dong2024rlhf} for one epoch with a batch size of 16. 

For our proposed Decomposed Reward Models (DRMs), we apply Principal Component Analysis (PCA) using scikit-learn's default settings to get the component vectors. We experiment with 100 heads for all methods. Note that DRMs utilize 50 distinct reward heads, and including their negative counterparts results in a total of 100 heads.

\subsection{Scalability of DRMs}
DRMs are already highly efficient and require significantly lower computational cost than training-based methods. In our experiments, we applied DRMs to a large preference dataset of size 0.5M, one of the largest open-source preference datasets. For comparison, training-based reward models using LLaMA3-8B (4096 hidden dimension) require 1-2 hours on a 48GB A6000 GPU. In contrast, the PCA step in DRMs runs in less than 1 minutes on the same server using only CPUs. 

While PCA involves covariance matrix computations that can become challenging at extreme scales, this is not a bottleneck for current dataset sizes (<1M). For larger datasets, scalable PCA solvers can be applied, eg. Incremental PCA~\cite{ross2008incremental}, Distributed PCA and Randomized PCA. These methods are well-supported in libraries like scikit-learn and PySpark, enabling DRMs to scale to millions of samples efficiently. To validate this, we conducted a scalability test using Incremental PCA across varying dataset sizes. As shown in Table~\ref{tab:inc_pca_time}, Incremental PCA exhibits approximately linear time complexity, demonstrating its practical scalability.

\begin{table}[ht]
\centering
\begin{tabular}{|l|c|c|c|}
\hline
\textbf{Dataset size} & \textbf{0.5 M} & \textbf{1 M} & \textbf{10 M} \\
\hline
Running Time  & 45.59 s & 97.09 s & 1117.13 s \\
\hline
\end{tabular}
\caption{Running time of Incremental PCA with different dataset sizes.}
\label{tab:inc_pca_time}
\end{table}

\end{document}